\documentclass[10pt,onecolumn,letterpaper]{article}
\usepackage{cvpr}
\usepackage{times}
\usepackage{epsfig}
\usepackage{graphicx}
\usepackage{amsthm}
\usepackage{amsmath}
\usepackage{amssymb}
\usepackage[breaklinks=true,bookmarks=false]{hyperref}

\usepackage[utf8]{inputenc} % allow utf-8 input
\usepackage[T1]{fontenc}    % use 8-bit T1 fonts
\usepackage{hyperref}       % hyperlinks
\usepackage{url}            % simple URL typesetting
\usepackage{booktabs}       % professional-quality tables
\usepackage{amsfonts}       % blackboard math symbols
\usepackage{nicefrac}       % compact symbols for 1/2, etc.
\usepackage{microtype}      % microtypography
\usepackage{xcolor}         % colors

\usepackage{multirow} 

\usepackage{soul}
\usepackage{url}

\usepackage{algorithm}
\usepackage{algorithmic}
\usepackage{epsfig}
\usepackage{epstopdf}
\usepackage{datetime}
\usepackage{diagbox} 
\usepackage{array}
\usepackage[T1]{fontenc}

\numberwithin{equation}{subsection}
\usepackage{datetime}
\usepackage[a4paper,margin=1in,headheight=15pt,headsep=24pt]{geometry}

\usepackage{fancyhdr}

\cvprfinalcopy % *** Uncomment this line for the final submission
\def\cvprPaperID{****} % *** Enter the CVPR Paper ID here
\allowdisplaybreaks

\begin{document}
\lhead{}
\lfoot{\date{\today},\date{\currenttime}}
\rfoot{NGD for DL}

\title{Gradient-based Fine-Tuning through Pre-trained Model Regularization}
\author{
% Xuanbo Liu, Liu Liu, Fuxiang Wu, Fusheng Hao, Xianglong Liu
\textbf{Xuanbo Liu}$^{1}$\quad
\textbf{Liu Liu}$^{1,2*}$\quad
\textbf{Fuxiang Wu}$^{3}$\quad
\textbf{Fusheng Hao}$^{3}$\quad
\textbf{Xianglong Liu}$^{1}$\\
[0.5em]
$^1$School of Artificial Intelligence, Beihang University\\
$^2$Hangzhou International Innovation Institute, Beihang University\\
$^3$Shenzhen Institutes of Advanced Technology, Chinese Academy of Sciences\\
}
\maketitle
\begingroup
\renewcommand\thefootnote{*}
\footnotetext{Corresponding author: \texttt{liuliubh@buaa.edu.cn}}
\endgroup

\begin{abstract}
Large pre-trained models have demonstrated extensive applications across various fields. However, fine-tuning these models for specific downstream tasks demands significant computational resources and storage. One fine-tuning method, gradient-based parameter selection (GPS), focuses on fine-tuning only the parameters with high gradients in each neuron, thereby reducing the number of training parameters. Nevertheless, this approach increases computational resource requirements and storage demands.
In this paper, we propose an efficient gradient-based and regularized fine-tuning method (GRFT) that updates the rows or columns of the weight matrix. We theoretically demonstrate that the rows or columns with the highest sum of squared gradients are optimal for updating. This strategy effectively reduces storage overhead and improves the efficiency of parameter selection. Additionally, we incorporate regularization to enhance knowledge transfer from the pre-trained model.
GRFT achieves state-of-the-art performance, surpassing existing methods such as GPS, Adapter Tuning, and LoRA. Notably, GRFT requires updating only 1.22\% and 0.30\% of the total parameters on FGVC and VTAB datasets, respectively, demonstrating its high efficiency and effectiveness. 
The source code will be released soon.

\end{abstract}

\section{Introduction}
The application of large models is becoming increasingly widespread, with concrete implementations across various industries. 
Multimodal dialogue interactions based on large models, such as GPT-4 \cite{openai2024gpt4technicalreport}, LlaMA \cite{touvron2023llama2openfoundation}, demonstrate that general-purpose large models can be effectively applied across multiple domains. 
In specific fields, models are typically trained using pre-trained models derived from large datasets. 
% Based on the pre-trained models, different datasets are selected for fine-tuning according to various downstream tasks, enabling adaptive application to specific domain tasks. 
% Pre-trained models are adapted through fine-tuning with task-specific datasets
Pre-trained models are fine-tuned with task-specific datasets tailored to downstream needs, thereby enhancing their applicability to domain-specific scenarios.
Fine-tuning involves further training a pre-trained large model on domain-specific data. This process optimizes the performance of the pre-trained model for a particular task, 
% enabling the model to perform specific downstream tasks effectively.
thereby enhancing its effectiveness in downstream applications.

Therefore, the application prospects of large models are extensive, spanning fields such as natural language processing, computer vision, and scientific research. However, due to the massive number of parameters, fine-tuning a scientifically accurate downstream task model requires substantial computational resources. 
Additionally, excessive memory consumption and high hardware requirements arise during training, limiting accessibility for researchers with limited resources \cite{li2024large}.
% Additionally, training these models incurs excessive memory consumption and imposes stringent hardware requirements.
Moreover, simply reducing the number of parameters in large models can degrade training accuracy, as it may compromise the model's ability to capture complex patterns in the data. These challenges present significant obstacles to the training of large-scale models,
rendering the trade-off between fine-tuning efficiency and model performance a critical area of research.

To tackle these issues,  various parameter-efficient fine-tuning (PEFT) methods have been proposed to reduce  computational costs while improving the performance. For instance, LoRA \cite{hu2021lora}, Adapter Tuning \cite{Houlsby_Giurgiu_Jastrzębski_Morrone_Laroussilhe_Gesmundo_Attariyan_Gelly_2019} and Vision Prompt Tuning \cite{jia2022visual}  introduce additional trainable parameters into the model's backbone or input space and fine-tune only these learnable parameters to improve efficiency. 
However, these methods introduce inference latency or break the original model structure, which results in increased inference time and a loss of the pre-trained model's expressiveness \cite{10688138}. Therefore, methods based on fine-tuning the model's own parameters have also been proposed. 
% One is Gradient-Based Parameter Selection (GPS) \cite{zhang2024gradient}, a new parameter-efficient fine-tuning approach, which selects a few parameters from the pre-trained model for tuning. 
One such approach is Gradient-Based Parameter Selection (GPS) \cite{zhang2024gradient}, a novel parameter-efficient fine-tuning method that selects a subset of parameters from the pre-trained model for tuning.
The method achieves outstanding performance. 
However, it has a disadvantage: the gradient-based selection method relies on the mask to freeze the parameters. 
The storage of masks incurs significant costs because the mask is sparse, which means that it has to store a mask matrix of the same size as the weight matrix. 
Additionally, 
% the sparse mask in training and updating is not good for hardware. 
the sparse nature of the masks poses challenges for hardware efficiency during training and updates.

In our proposed method, we introduce gradient-based fine-tuning with pre-trained model regularization for parameter-efficient fine-tuning. In particular, rather than selecting sparse parameters, we select entire rows or columns of the weight matrix, meaning that only the indices of these rows or columns need to be stored in the mask. This approach significantly reduces storage costs, while also simplifying the masking mechanism, making it more efficient in terms of memory usage and computational overhead. Additionally, by selecting structured groups of parameters, the method aligns better with modern hardware optimizations, facilitating improved performance during training and inference.
Additionally, to enhance knowledge transfer from the pre-trained model and improve accuracy, we incorporate an $L_2$ regularization term into the loss function. 
Besides, to filter high-quality datasets for better selection of effective parameters, we investigate data processing techniques that compute the mask associated with the minimal loss by randomly partitioning the training data into several subsets. 
We hypothesize that a mask corresponding to minimal loss reflects higher-quality data, which offers the advantage of more effective mask determination. 
To evaluate our method, we conduct experiments and evaluated our method on image classification tasks and text classification tasks. Our proposed method achieves state-of-the-art performance compared to GPS and other PEFT methods while using fewer parameters in certain tasks. 

Overall, our contributions are summarized as follows:
\begin{itemize}
    \item We propose new gradient-based parameter selection frameworks, GRFT, 
    a Gradient-based and Regularized Fine-Tuning method that only trains the parameters associated with large gradients and additional regularization constraints.
    \item We introduce a novel gradient-based parameter selection method to reduce storage requirements, which fine-tune the  entire rows or columns of the parameter matrix. Moreover, a theoretical justification is provided for selecting parameters with larger squared gradients. Additionally, we introduce regularization constraints that limit parameters sizes to be close to those of the pre-trained parameters, thereby facilitating knowledge transfer and enhancing generalization.
    \item Empirical evaluations over image classification and text classification across ViT models and LLaMA-3 models demonstrate that  our method outperforms fine-tuning methods such as GPS and LoRA in terms of accuracy, while not significantly increasing the parameter count.
\end{itemize}

\section{Related works}
\noindent\textbf{Parameter-Efficient Fine Tuning} 
Parameter-efficient fine-tuning is a widely used fine-tuning method in both computer vision and natural language processing, focusing on training parts of the model parameters or fine-tuning additional modules, which has the advantages of lower computational cost and shorter time requirements compared to full fine-tuning. Lately, various existing techniques, including Adapter Tuning \cite{karimi2021compacter,lu2023uniadapter,zhang2023llama}, Prompt Tuning \cite{jia2022visual,zhou2022learning,wang2023lion}, LoRA \cite{hu2021lora} and its variants \cite{qiang2024bilora,hayou2024lora+}, are attempting to maintain the model performance while reducing the computation and storage costs. In a recent study, GPS \cite{zhang2024gradient} fine-tune a few parameters from the pre-trained model while freezing the reminder of the model. The selection of these parameters depends on their individual gradients. The advantage of this method is that it does not introduce additional computational costs and parameters, and it has good adaptability to any agnostic models.

\noindent\textbf{Transfer Learning and Regularization}
Fine-tuning is essentially a transfer learning strategy that leverages the knowledge that the model has learned from large-scale datasets, enabling the model to be fine-tuned with a smaller, task-specific dataset. The key advantage of fine-tuning is that it allows the model to converge faster and achieve better performance with less data compared to training from scratch. However, excessive fine-tuning can lead to catastrophic forgetting. 
% This phenomenon occurs when the model's parameters are overly updated during the fine-tuning process, causing the model to forget the knowledge it had previously learned from the pre-training phase. 
As a result, the model may experience a decline in performance on the tasks \cite{toneva2018empirical,FRENCH1999128}. 
Regularization can improve the generalization ability of models, such as Ridge Regression \cite{Hoerl2000RidgeRB}. It introduces a regularization term proportional to the square of the magnitude of the model parameters. Additionally, Ridge Regression \cite{Hoerl2000RidgeRB} improves the generalization ability of models by preventing overfitting. The $L_2$ regularization provided by Ridge Regression helps reduce this risk by penalizing excessively large weights.
% This leads to a more robust model that can perform well not only on the fine-tuning task but also on other tasks that may be related.
In our methods, we add $L_2$ regularization in the training loss function to limit the parameters close to pre-trained parameters, achieveing knowledge transferring and regularization. 

\noindent\textbf{Sub-network training} 
Research have confirmed the effectiveness of sub-network training. Fine-tuning a portion of the model parameters can also achieve satisfactory performance. \cite{Ouyang_Wang_Zhang_Yang_2016} investigates fine-tuning only the last few layers of deep neural network and find that it could reduce the training time and computation costs while keeping a comparable performance to full fine-tuning. And the lottery ticket hypothesis \cite{Frankle_Carbin_2018} indicates the sub-networks can achieve the same accuracy of the entire model. Pruning techniques \cite{Gale_Elsen_Hooker_2019,Han_Mao_Dally_2015,Han_Pool_Tran_Dally_2015,Kruschke_Movellan_1991,Li_Kadav_Durdanovic_Samet_Graf_2016,Wen_Wu_Wang_Chen_Li_2016,kurtic2023ziplminferenceawarestructuredpruning} also underscore the importance of sub-networks in model efficiency. However, these approaches are not aligned with the PEFT paradigm. In contrast, we focus on freezing a number of parameters and tuning only small parts to fit the downstream task, thereby achieving efficient fine-tuning in the PEFT setting.

\section{Proposed Method}
In this section, we introduce the gradient-based fine-tuning approach, a parameter-efficient fine-tuning method, that focuses on selecting and fine-tuning the parameters with large gradients.
This approach is based on the principle that parameters with larger gradients undergo more significant updates during training, highlighting the critical role
in downstream tasks. 
The method primarily comprises the following two key components: gradient selection and model regularization.

\subsection{Gradient Selection}
In the gradient-based parameter selection methods, the most crucial step is determining how to effectively select gradient.  Authors \cite{zhang2024gradient} consider that each neuron has several vital input parts, which contribute most in the training of downstream tasks. 
Accordingly, they select the top gradients from each neuron. 
With the approach, it is possible to make each neuron contribute to the downstream tasks.
However, the gradient computation for each neuron imposes significant hardware requirements. 
Currently, existing hardware struggles to compute gradients solely for the sparse parts of the weight matrix. 
Therefore, in practical implementations, the gradient of the entire weight matrix is still computed, with the corresponding gradients masked during the update process. 
What's more, the mask has the same dimensions as the weight matrix, which leads to the increased  storage overhead. 

\begin{figure*}[t] 
    \centering
    \includegraphics[width=\textwidth]{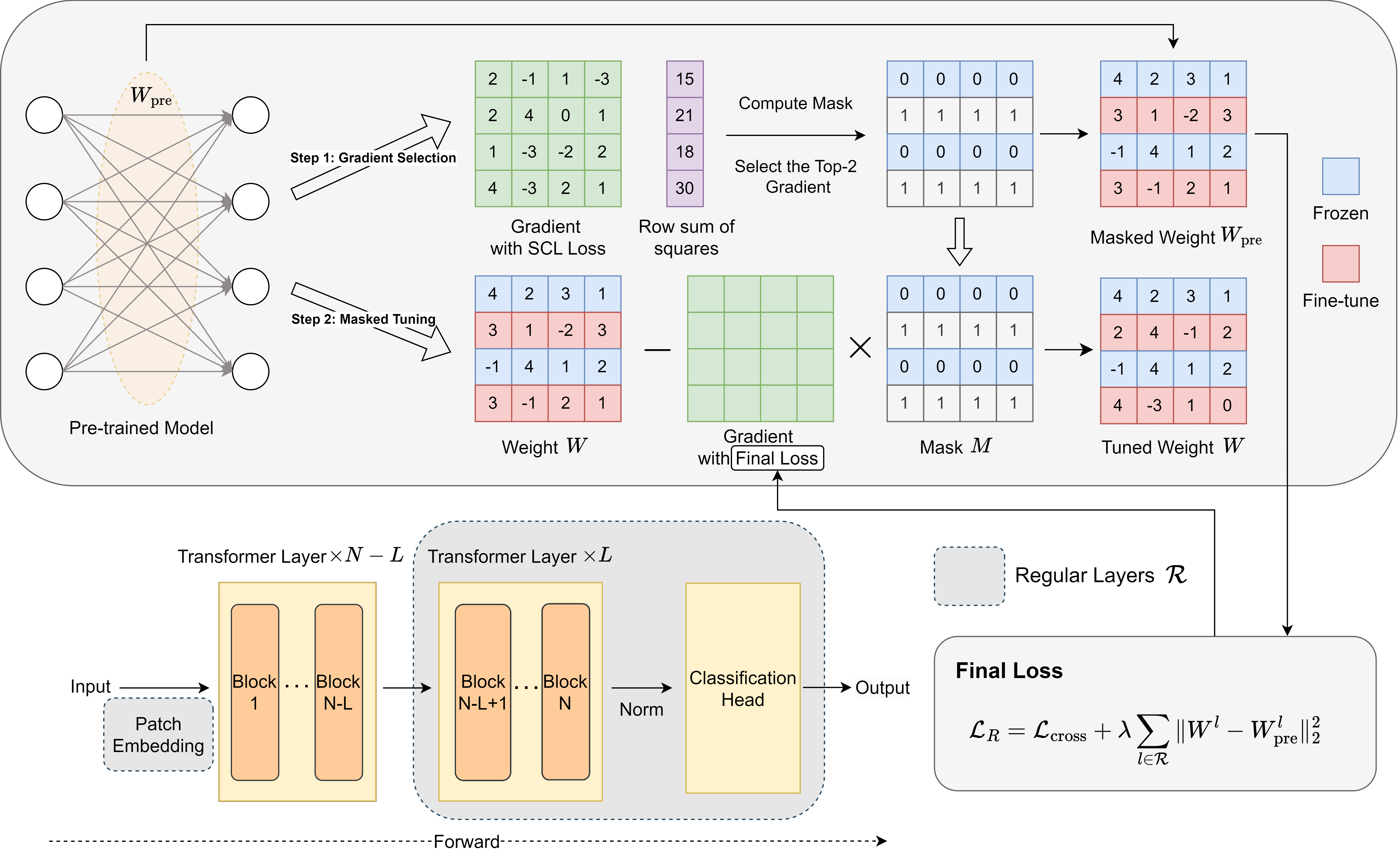} 
    \caption{The overall pipeline of our approach. This approach primarily consists of two steps.
    Step 1: Gradients Selection. 
    Before training, we firstly compute gradients of the pre-trained model with SCL loss and calculate squares sum of the each row. Then,we gain the mask to freezing the parameters. 
    Step 2: Masked Tuning. 
    % In training update, we have the update equation list above. 
    For pre-trained models, an additional constraint is incorporated alongside the traditional cross-entropy loss function to account for knowledge transfer from the pre-trained models.
} 
    \label{fig:method} 
\end{figure*}

\textbf{The Principle of Gradient Selection}
To address this issue, we adopt a row-wise selection strategy. In the strategy, for the weight matrix $W$, we only select the row parameters with large gradient. Compared to sparse matrices, selecting gradient parameters row by row imposes lower hardware demands and is more beneficial for subsequent applications and practical implementations. Furthermore, the storage cost is reduced since the mask only needs to store the indices of the rows, rather than the entire mask.

% \subsubsection{Principles and Selection methods}
At the time step $t$ with the model parameters $\theta ^t$, we consider $\mathcal{L}$ to be the loss function. The first-order Taylor expansion is
% \begin{align}
%     \mathcal{L}(x;\theta^{t+1}) \approx \mathcal{L}(x;\theta^{t})+<\nabla \mathcal{L}(x;\theta^{t}), \theta^{t+1}-\theta^{t}>.\label{loss expansion}
% \end{align}
\begin{align}
    \mathcal{L}(\theta^{t+1}) \approx \mathcal{L}(\theta^{t})+\langle\nabla \mathcal{L}(\theta^{t}), \theta^{t+1}-\theta^{t}
    \rangle.\label{loss expansion}
\end{align}
At the same time, if we assume that the learning rate is $\alpha_t$ at time step $t$, it is evident that for gradient descent optimization, there is $\theta^{t+1}-\theta^{t}=
-\alpha_t \nabla \mathcal{L}(\theta^t) \odot M$. Here $\odot$ is element-wise multiplication and $M$ is the gradient mask to freeze the parameters. Put it in the above Eq. \eqref{loss expansion}: 
\begin{align}
    \Delta \mathcal{L}=\mathcal{L}(\theta^{t+1}) - \mathcal{L}(\theta^{t}) \nonumber 
    \approx - \langle\nabla \mathcal{L}(\theta^{t}),\alpha_t\nabla \mathcal{L}(\theta^t)\odot M \rangle.
\end{align}
We determine $M$ to facilitate a larger decrease in the loss during each iteration:
\begin{align}
    \label{Eq:principleSelection1} M &=\text{argmax}_M \langle\nabla \mathcal{L}(\theta^{t}), \nabla \mathcal{L}(\theta^t) \odot M \rangle \\ 
    \label{Eq:principleSelection2} &=\text{argmax}_M \langle\nabla \mathcal{L}(\theta^{t})\odot M, \nabla \mathcal{L}(\theta^t)\odot M \rangle. 
\end{align}

Note that since $M$ is a binary matrix with values of 0 or 1, when $\nabla \mathcal{L}(\theta^t) = \lbrack \nabla \mathcal{L}(\theta^t)_1, \nabla \mathcal{L}(\theta^t)_2, \cdots, \nabla \mathcal{L}(\theta^t)_d \rbrack \in \mathbb{R}^d$ is a one-dimensional vector and the mask $M = \lbrack M_1,M_2,\cdots,M_d \rbrack \in \mathbb{R}^d$, performing the element-wise multiplication with $M$ results in zeros at the corresponding positions. We have 
\begin{align}
    \langle\nabla \mathcal{L}(\theta^{t}), \nabla \mathcal{L}(\theta^t) \odot M \rangle =&\sum^d_{i=1} \nabla \mathcal{L}(\theta^{t})_i \cdot (\nabla \mathcal{L}(\theta^t)_i \cdot M_i ) \nonumber \\ 
    =& \sum^d_{i=1} (\nabla \mathcal{L}(\theta^{t})_i \cdot M_i )^2 
    \nonumber\\ 
    =&\langle\nabla \mathcal{L}(\theta^{t})\odot M, \nabla \mathcal{L}(\theta^t)\odot M \rangle,  \nonumber
\end{align}
Therefore, when computing Eq. \eqref{Eq:principleSelection1}, the values at those positions in the final result will also be zero. This leads to the conclusion that Eq. \eqref{Eq:principleSelection2} holds. 

\textbf{The Computation of Gradient Mask}
Accordingly, we can infer that if $M$ is sparse, the Eq. \eqref{Eq:principleSelection2} implies a preference for retaining the largest gradients in the entire weight matrix. Therefore, it is reasonable for GPS \cite{zhang2024gradient} to select based on neurons, retaining the largest value in each row of the gradient matrix. However, the sparse gradient computation has vital hardware requirements. And the mask is stored as the same size of weight matrix, making it a high storage cost.
Therefore, our proposed approach is to select the entire row or column, which is theoretically justified to select the rows with the largest squared sums of the gradients.
Given the pretrained parameter $W_\text{pre}$,  we obtain the optimal fine-tuned parameter based on with gradient $\nabla \mathcal{L}^\text{scl}(W_\text{pre})$ in the following optimization objective,
\begin{align}
    \text{min}_M&\|\nabla \mathcal{L}^\text{scl}(W_\text{pre})-\nabla \mathcal{L}^\text{scl}(W_\text{pre})\odot M\|_2^2 
    \\
    s.t. &\|M_{,j}\|_0\leq k, \quad \forall j\in[n] .\nonumber
\end{align}

% , Like GPS \cite{zhang2024gradient}, 

Note that, to prevent the effects of the randomly initialized classification head on fine-tuning, the SCL loss $\mathcal{L}^\text{scl}$ \cite{Khosla_Teterwak_Wang_Sarna_Tian_Isola_Maschinot_Liu_Krishnan_2020,zhang2024gradient} is used to calculate the gradients. The loss function is 
\begin{align}
    \mathcal{L}^{\text{scl}} = \sum_{i \in \mathcal{D}} \mathcal{L}^{\text{scl}}_i \nonumber  
    = \sum_{i \in \mathcal{D}} \frac{-1}{|\mathcal{P}(i)|} \sum_{p \in \mathcal{P}(i)} \log \frac{\exp(z_i \cdot z_p / \tau)}{\sum_{a \in \mathcal{A}(i)} \exp(z_i \cdot z_a / \tau)},
\end{align}
where $\mathcal{D}$ is train dataset, 
$i$ denotes the $i_{\text{th}}$ sample in $\mathcal{D}$; 
$\mathcal{P}(i)$ represents a subset of $\mathcal{D}$, which includes all samples that belong to the same class as $i$; $\mathcal{A}(i)$ defines the set of all samples in $\mathcal{D}$ excluding $i$; $z$ represents the feature and $\tau \in \mathbb{R}^+$ is a scalar temperature parameter, `$\cdot$' is the inner (dot) product.

Based on the conclusion in Eq. \eqref{Eq:principleSelection2}, 
we compute the sum of the squared gradient of each row. Let the squared sum of the $i_\text{th}$ row be defined as: 
\begin{align}
    S_i = \sum_{j=1}^n h_{ij}^2, \quad \text{for } i = 1, 2, \dots, m,
\end{align}
where $h_{ij}$ is the element of $H\equiv \nabla \mathcal{L}^\text{scl}(W)$ in the $i_\text{th}$ row and $j_\text{th}$ column. We then select the indices corresponding to the top-$k$ largest values in $\{S_1, S_2, \dots, S_m\}$, denoted the selected index set as $\mathcal{T}$. A mask $M$ is constructed as follows:
\begin{align}
    M_{ij} = 
    \begin{cases} 
    1, & \text{if } i \in \mathcal{T}, \\
    0, & \text{otherwise}.
    \end{cases}
    \label{Eq:maskSelection}
\end{align}
Therefore, the mask determines the parameters we select. In the backpropagation update process, only the selected parameters are updated. For the $t$ training epoch, the $l_\text{th}$ layer of the model has the following:
\begin{align}
    W^l_{t+1}-W^l_{t}=-\eta_t \nabla \mathcal{L}(W^l_t) \odot M^l,
\end{align}
where $W^l_{t}$ means the parameters of $l$-$\text{th}$ layer at $t$ step and $M^l$ is the mask of $l$-$\text{th}$ layer.

\subsection{Model Regularization}
The core of fine-tuning large models lies in transferring and applying the vast amount of knowledge the models have acquired during pre-training to specific downstream tasks. Pre-trained models are trained on massive datasets and learn a widely general features, representations, and patterns that can be very useful for various applications. When fine-tuning for a new task, it is essential to ensure that the knowledge gained during pre-training is effectively preserved and adapted to the new context.

One of the key challenges in fine-tuning is preventing the model from losing its generalizable knowledge while adapting to the new task. This phenomenon, known as "catastrophic forgetting", occurs when the model overly adjusts its parameters in response to the new task, effectively "forgetting" the valuable information from pre-training \cite{toneva2018empirical,FRENCH1999128}. To mitigate this issue, various regularization techniques can be employed during the fine-tuning process. We employ an $L_2$ norm constraint to the loss function during training, limiting the model parameters to a smaller range during the fine-tuning process and facilitating the transfer of knowledge from the pre-trained model to the downstream task.
Moreover, as parameters of the classification head are initialized using a uniform distribution within the interval $[- \sqrt{k}, \sqrt{k}]$, where $k = \frac{1}{d}$ and $d$ is the input feature dimension \cite{he2015delving}, the interval is small and centered around zero due to the large input feature dimension. Therefore, applying the $L_2$ norm to the classification head can reduce the complexity of the classification head's parameters. 
% and serves as a regularization term in the optimization objective function. 
This helps prevent the model from becoming overly complex in order to fit the training set, thus reducing the risk of overfitting and improving the model's generalization ability. The overall objective function is
\begin{align}
\mathcal{L}_R=\mathcal{L}_\text{cross}+\lambda \|W-W_\text{pre}\|_2^2.
    \label{Eq. overall objective loss function}
\end{align}
where $\mathcal{L}_R$ is the final loss function, $\mathcal{L_{\text{cross}}}$ is the cross-entropy loss function during the training process, $\lambda$ is the regularization parameter for the $L_2$ norm. 
However, since the parameters of the last few layers of the model have a significant impact on the training results during fine-tuning \cite{zhang2023llama}, the constraint primarily targets these layers. The final loss function includes both the original loss function and the modified regularization function, specifically as follows:
\begin{align}
\mathcal{L}_R=\mathcal{L}_\text{cross}+\lambda \sum_{l \in \mathcal{R}}\|W^l-W_\text{pre}^l\|_2^2,
    \label{Eq. final loss function}
\end{align}
where $\mathcal{R}$ is the regular layers set consisting of the last $L$ layers, patch embedding, and classification head. 
% We make the ablation study regard to the section of constraint layers sets.

% \begin{center}
% \begin{minipage}{.7\linewidth}
\begin{algorithm}[t]
\caption{Gradient-based and Regularized Fine-tuning (GRFT)}
\label{alg:GRFT}
\textbf{Input:} A layer weight matrix $W \in \mathbb{R}^{m \times n}$. Epochs $N$, learning rate $\eta_t$, decay rates $\beta_1$, $\beta_2$, select row number $k$, scale hyperparameter $\lambda$, regular module set $\mathcal{R}$.
% \STATE Initialize $t \leftarrow 0$
\begin{algorithmic}[1]
\STATE \textbf{Computing mask} $M$
    \STATE \hspace{1em} $H  \leftarrow \nabla \mathcal{L}^{\text{scl}}(W_\text{pre})\in \mathbb{R}^{m \times n}$
    \STATE \hspace{1em} $S_i$ $\leftarrow$ $h_{i1}^2+h_{i2}^2+\cdots+h_{in}^2$,$i=1,2,\cdots,m$
    \STATE \hspace{1em} Sort $\mathcal{S}=\{S_1, S_2, \dots, S_m\}$ 
    % \STATE \hspace{1em} $T \leftarrow $the top-$k$ largest values indices in $S$
    % \STATE \hspace{1em} $m_{ij} \leftarrow 1$ if $i\in T$ else $0$
    \STATE \hspace{1em} Obtain the mask $M$ based on Eq. \eqref{Eq:maskSelection}
\STATE \textbf{Training}
\FOR {$t = 1$ to $N$}
    \STATE $\mathcal{L}_R=\mathcal{L}_\text{cross}+\lambda \sum_{l \in \mathcal{R}} ||W^l_t-W^l_0||_2^2$
    \STATE \textbf{UPDATE} $(\hat{g}_t)$ \textbf{by Adam}:
    \STATE $\hat{g}_t=\nabla \mathcal{L}_R(W_t) \odot M$
    \STATE $m_t \leftarrow \beta_1 m_{t-1} + (1 - \beta_1) \hat{g}_t$
    \STATE $v_t \leftarrow \beta_2 v_{t-1} + (1 - \beta_2) \hat{g}_t^2$
    \STATE $\hat{m}_t \leftarrow m_t / (1 - \beta_1^t)$
    \STATE $\hat{v}_t \leftarrow v_t / (1 - \beta_2^t)$
    \STATE $W_t \leftarrow W_{t-1} - \eta_t \hat{m}_t / \sqrt{\hat{v}_t + \epsilon}$
    \STATE $t \leftarrow t + 1$
    \ENDFOR
\end{algorithmic}
\end{algorithm}
% \end{minipage}
% \end{center}

\textbf{Gradient-based and Regularized Fine-tuning Algorithm}
% Recent research highlights the critical importance of high-quality datasets in the fine-tuning of machine learning models. A key observation is that the quality of the data directly influences the gradients we select during the training process. Specifically, the gradients are related to the data used for model optimization, and the choice of data subset can significantly impact the performance of the model. In our approach, we randomly partition the original training dataset into several subsets. Each subset is independently used for model training to compute the loss and gradients. The gradient associated with minimal loss is considered optimal. This suggests that the data subset that generates this gradient holds superior quality, as it leads to more effective model optimization. A high-quality subset provides more informative or less noisy examples, resulting in more reliable gradient updates. Consequently, such a subset is more likely to contribute to better model generalization and overall performance. Through this methodology, we can refine our understanding of how different portions of data impact model learning, offering a novel approach to identifying and leveraging high-quality data for fine-tuning.
We present the proposed method in Algorithm \ref{alg:GRFT}, which consists of two main parts:  Mask Computation and Training. Specifically, before training begins, we first compute the model gradients under the SCL loss $\nabla \mathcal{L}^{\text{scl}}(W_\text{pre})$. Subsequently, we calculate the sum of squares for each row or column of the gradient $S_i$ and select the top $k$ rows or columns with the largest sums. Based on Eq. \eqref{Eq:maskSelection}, we compute our mask $M$. During the training phase, for the standard loss function (such as cross-entropy $\mathcal{L}_\text{cross}$), we add an $L_2$ regularization term, thereby obtaining the final loss function $\mathcal{L}_R$ as defined in Eq. \eqref{Eq. final loss function}. In this process, we employ the Adam \cite{Kingma_Ba_2014} optimizer, where we set $\hat{g}$ in Adam to be the masked gradient and then the standard Adam update procedure is applied to iteratively update the weight matrix until convergence.
% proceed with the normal Adam update procedure to update our weight matrix until the training ended.

\section{Experiments}
\subsection{Implementation}
In image classification tasks, the model we implement is vit-base-patch16-224-in21k \cite{Dosovitskiy_Beyer_Kolesnikov_Weissenborn_Zhai_Unterthiner_Dehghani_Minderer_Heigold_Gelly_etal._2020}. The model uses $16 \times 16$ image patches as inputs and is pre-trained on ImageNet-21k \cite{Deng_Dong_Socher_Li_Kai_-Li_Li_Fei-Fei_2009} at resolution $224 \times 224$. By pretraining on the dataset, the model learns the internal representation of the images, which can be used in downstream tasks to extract features. 
We use the Adam \cite{Kingma_Ba_2014} optimizer and apply a cosine learning rate decay for fine-tuning. Each downstream task is trained for 100 iterations, with an additional 10 warm-up epochs for the learning rate before the training iterations begin. All experiments are conducted on an NVIDIA GeForce RTX 4090 GPU.
In text classification tasks, we implement Llama3.2 1B model \cite{grattafiori2024llama3herdmodels} and fine-tuning in CoLA, MRPC and RTE datasets of GLUE benchmark \cite{wang2019gluemultitaskbenchmarkanalysis}. We added a linear classification head module to the model to perform our classification task. The optimizer we use is AdamW \cite{loshchilov2019decoupledweightdecayregularization}. The experiments are conducted on an NVIDIA A800 80GB PCIe GPU.

\subsection{Datasets}
% \begin{itemize}
    % \item 
    \noindent\textbf{FGVC (Fine-Grained Visual Classification)}: FGVC is a subset of image classification tasks which mainly deal with distinguishing between visually similar objects within a category. FGVC datasets include:
        % \begin{itemize}
        % \item 
        Stanford Dogs \cite{KhoslaYaoJayadevaprakashFeiFei_FGVC2011}: A dataset includes 120 breeds of dogs with 20,580 images. 
        % \item 
        Stanford Cars \cite{6755945}: A dataset consists of 16,185 images from 196 classes of cars. The training split comprises 8,144 images and the test comprises 8,041 images.
        % \item 
        Nabirds \cite{Horn_2015_CVPR}: A dataset contains 102 flower species with 8,189 images.
        % \item 
        CUB\_200\_2011 \cite{Wah2011TheCB}: A dataset for fine-grained bird species classification containing 11,788 images of 200 bird species. 
        % \item 
        Oxfordflower 102 \cite{Nilsback08}: A dataset consists of 102 flower categories. The number of images in each class is between 40 and 258.
        % \end{itemize}
    % \item

    \noindent\textbf{VTAB (Visual Task Adaptation Benchmark)}: a benchmark designed to evaluate the performance of transfer learning techniques in visual tasks, testing the performance of models trained on one set of tasks generalizing to a wide variety of other visual tasks. VTAB includes 19 different datasets, covering various visual domains. 
        % \begin{itemize}
        %     \item 
            Natural: includes tasks like CIFAR-100 \cite{krizhevsky2009learning} and Caltech101 \cite{1384978}.
            % \item 
            Specialized: includes tasks like Patch Camelyon \cite{Veeling2018-qh} and Resisc45 \cite{Cheng_2017}. 
            % \item 
            Structured: includes tasks like DMLab \cite{zhai2020largescalestudyrepresentationlearning} and Clevr \cite{johnson2017clevr}.
        % \end{itemize}

    \noindent\textbf{GLUE (General Language Understanding Evaluation)} \cite{wang2019gluemultitaskbenchmarkanalysis}: a benchmark dataset designed to measure the capabilities of models in natural language understanding (NLU). It consists of various subtasks, including text classification, sentence similarity evaluation, and natural language inference (NLI), among others. GLUE is primarily used to assess the performance of pre-trained language models.
% \end{itemize}

\subsection{Experimental Results}
We compare with different fine-tuning methods, including full fine-tuning, linear and bias \cite{zaken2022bitfitsimpleparameterefficientfinetuning}, Adapter \cite{houlsby2019parameterefficienttransferlearningnlp}, LoRA \cite{hu2021lora}, VPT \cite{jia2022visual}, SSF \cite{lian2023scalingshiftingfeatures}, SPT \cite{he2023sensitivityawarevisualparameterefficientfinetuning} and GPS \cite{zhang2024gradient}. Except for GPS, the results of the other methods follow the results in the GPS paper, while the GPS results are reproduced by ourselves, marked as GPS*. Our results are shown in Table \ref{Table:FGVC performances} and Table \ref{Table:VTAB performances}.
% Table generated by Excel2LaTeX from sheet 'Sheet2'

\begin{table}[t]
  \renewcommand{\arraystretch}{1.5}
  \centering
  \caption{Comparisons results on FGVC with ViT-B/16 models pre-trained on ImageNet-21K.}
  \resizebox{0.8\textwidth}{!}{
    \begin{tabular}{l|ccccc|cc}
    \toprule
    \multicolumn{1}{c|}{Dataset} &
    \multicolumn{1}{c}{\centering CUB-2011} &
    \multicolumn{1}{c}{\centering NABirds} &
    \multicolumn{1}{c}{\centering Oxford Flowers} &
    \multicolumn{1}{c}{\centering Stan.Dogs} &
    \multicolumn{1}{c|}{\centering Stan.Cars} & 
    \multicolumn{1}{c}{\centering Mean Acc.} &
    \multicolumn{1}{c}{\centering Params.(\%)}\\
    \midrule
    Full  & 87.3  & 82.7  & 98.8  & 89.4  & 84.5  & 89.44  & 100.00  \\
    \midrule
    Linear & 85.3  & 75.9   & 97.9  & 86.2  & 51.3  & 79.32  & 0.21  \\
    Bias \cite{zaken2022bitfitsimpleparameterefficientfinetuning}  & 88.4  & 84.2  & 98.8  & \underline{91.2}  & 79.4  & 88.40  & 0.33  \\
    \midrule
    Adapter \cite{houlsby2019parameterefficienttransferlearningnlp} & 87.1  & 84.3  & 98.5  & 89.8  & 68.6  & 85.66  & 0.48  \\
    LoRA \cite{hu2021lora}  & 85.6  & 79.8  & 98.9  & 87.6  & 72.0  & 84.78  & 0.90  \\
    VPT-Shallow \cite{jia2022visual} & 86.7  & 78.8  & 98.4  & 90.7  & 68.7  & 84.66  & 0.29  \\
    VPT-Deep \cite{jia2022visual} & 88.5  & 84.2  & 99.0  & 90.2  & 83.6  & 89.10  & 0.99  \\
    SSF \cite{lian2023scalingshiftingfeatures}   & 89.5  & 85.7  & 99.6  & 89.6  & 89.2  & 90.72  & 0.45  \\
    SPT-Adapter\cite{he2023sensitivityawarevisualparameterefficientfinetuning} & 89.1  & 83.3  & 99.2  & 91.1  & 86.2  & 89.78  & 0.47  \\
    SPT-LoRA \cite{he2023sensitivityawarevisualparameterefficientfinetuning} & 88.6  & 83.4  & 99.5  & \textbf{91.4}  & 87.3  & 90.04  & 0.60  \\
    \midrule
    GPS* \cite{zhang2024gradient} & \underline{89.6}  & \underline{86.8}  & \underline{99.7}  & 88.9  & \underline{90.4}  & \underline{91.06}  & 1.07 \\
    \midrule
    GRFT (\textbf{ours})  & \textbf{90.1}  & \textbf{87.0}  & \textbf{99.7}  & 89.1  & \textbf{90.8}  & \textbf{91.33}  & 1.22  \\
    \bottomrule
    \end{tabular}}%
  \label{Table:FGVC performances}%
\end{table}%

In Table \ref{Table:FGVC performances}, The average accuracy of GRFT is the highest, reaching 91.33\%, indicating that it has better generalization ability when handling different datasets. In contrast, the average accuracies of the other models range from 84.66\% (VPT-Shallow) to 91.06\% (GPS*), all of which are lower than GRFT. In terms of the number of parameters, GRFT has 1.22\% of the total parameters, which is considered above average among all methods. This suggests that GRFT achieves a high accuracy in the large datasets while increasing the number of parameters to be updated. Some other methods, although having fewer parameters, also show relatively lower accuracy. From this, it can be concluded that GRFT performs exceptionally well in the FGVC experiment.

\begin{table}[t]
  \renewcommand{\arraystretch}{1.5}
  \centering
  \caption{Comparisons results on VTAB-1k with ViT-B/16 models pre-trained on ImageNet-21K.}
  \resizebox{\textwidth}{!}{
    \begin{tabular}{l|ccccccc|cccc|cccccccc|cc}
    \toprule
    \multicolumn{1}{c|}{\multirow{6}{*}{\diagbox{Dataset}{Method}}} & \multicolumn{7}{c}{Natural} & \multicolumn{4}{c}{Specialized} & \multicolumn{8}{c}{Structured} & \multicolumn{2}{c}{VTAB}\\
\cline{2-22} 
& \rotatebox{90}{CIFAR-100} & \rotatebox{90}{Caltech101} & \rotatebox{90}{DTD}  & \rotatebox{90}{Flowers102}  & \rotatebox{90}{Pets}  & \rotatebox{90}{SVHN}  & \rotatebox{90}{Sun397}  & \rotatebox{90}{Patch Camelyon}  & \rotatebox{90}{EuroSAT}  & \rotatebox{90}{Resisc45}  & \rotatebox{90}{Retinopathy}  & \rotatebox{90}{Clevr/count}  & \rotatebox{90}{Clevr/distance}  & \rotatebox{90}{DMLab}  & \rotatebox{90}{KITTI/distance}  & \rotatebox{90}{dSprites/loc}  & \rotatebox{90}{dSprites/ori}  & \rotatebox{90}{SmallNORB/azi}  & \rotatebox{90}{SmallNORB/ele}  & \rotatebox{90}{Mean Acc.}  & \rotatebox{90}{Mean Params. (\%)} \\
    \midrule
    Full  & 68.9  & 87.7  & 64.3  & 97.2  & 86.9  & 87.4  & 38.8  & 79.7  & 95.7  & 84.2  & 73.9  & 56.3  & 58.6  & 41.7  & 65.5  & 57.5  & 46.7  & 25.7  & 29.1  & 65.57  & 100.00\\
    \midrule
    Linear & 63.4  & 85.0  & 64.3  & 97.0  & 86.3  & 36.6  & 51.0  & 78.5  & 87.5  & 68.6  & 74.0  & 34.3  & 30.6  & 33.2  & 55.4  & 12.5  & 20.0  & 9.6   & 19.2  & 53.00  & 0.05\\
    Bias \cite{zaken2022bitfitsimpleparameterefficientfinetuning}  & 72.8  & 87.0  & 59.2  & 97.5  & 85.3  & 59.9  & 51.4  & 78.7  & 91.6  & 72.9  & 69.8  & 61.5  & 55.6  & 32.4  & 55.9  & 66.6  & 40.0  & 15.7  & 25.1  & 62.05  & 0.16 \\
    \midrule
    Adapter \cite{houlsby2019parameterefficienttransferlearningnlp} & \underline{74.1} & 86.1  & 63.2  & 97.7  & 87.0  & 34.6  & 50.8  & 76.3  & 88.0  & 73.1  & 70.5  & 45.7  & 37.4  & 31.2  & 53.2  & 30.3  & 25.4  & 13.8  & 22.1  & 55.82  & 0.31\\
    LoRA \cite{hu2021lora}  & 68.1  & 91.4  & 69.8  & 99.0  & 90.5  & 86.4  & 53.1  & 85.1  & 95.8  & 84.7  & 74.2  & \underline{83.0}  & 66.9  & 50.4  & \underline{81.4}  & 80.2  & 46.6  & 32.2  & 41.1  & 72.63  & 0.90  \\
    VPT-Shallow \cite{jia2022visual} & 77.7  & 86.9  & 62.6  & 97.5  & 87.3  & 74.5  & 51.2  & 78.2  & 92.0  & 75.6  & 72.9  & 50.5  & 58.6  & 40.5  & 67.1  & 68.7  & 36.1  & 20.2  & 34.1  & 64.85  & 0.13  \\
    VPT-Deep \cite{jia2022visual} & \textbf{78.8} & 90.8  & 65.8  & 98.0  & 88.3  & 78.1  & 49.6  & 81.8  & 96.1  & 83.4  & 68.4  & 68.5  & 60.0  & 46.5  & 72.8  & 73.6  & 47.9  & \underline{32.9}  & 37.8  & 69.43  & 0.70  \\
    SSF \cite{lian2023scalingshiftingfeatures}  & 69.0  & 92.6  & \underline{75.1}  & \underline{99.4} & \textbf{91.8} & \underline{90.2}  & 52.9  & \underline{87.4}  & 95.9  & \textbf{87.4} & 75.5  & 75.9  & 62.3  & 53.3  & 80.6  & 77.3  & \underline{54.9} & 29.5  & 37.9  & 73.10  & 0.28  \\
    SPT-ADAPTER \cite{he2023sensitivityawarevisualparameterefficientfinetuning} & 72.9  & 93.2  & 72.5  & 99.3  & 91.4  & 88.8  & \textbf{55.8} & 86.2  & \underline{96.1}  & 85.5  & 75.5  & \underline{83.0}  & \textbf{68.0} & 51.9  & 81.2  & 51.9  & 31.7  & \textbf{41.2} & \textbf{61.4} & 73.03  & 0.44  \\
    SPT-LoRA \cite{he2023sensitivityawarevisualparameterefficientfinetuning} & 73.5  & \underline{93.3}  & 72.5  & 99.3  & \underline{91.5}  & 87.9  & \underline{55.5}  & 85.7  & \textbf{96.2} & 85.9  & 75.9  & \textbf{84.4} & \underline{67.6}  & 52.5  & \textbf{82.0} & \underline{81.0}  & 51.1  & 30.2  & 41.3  & \underline{74.07} & 0.63 \\
    \midrule
    GPS* \cite{zhang2024gradient}& 68.7  & \textbf{93.6} & 72.6  & 99.3  & 90.0  & 90.1  & 52.4  & 87.0  & 95.9  & \underline{86.5}  & \underline{76.1}  & 78.9  & 62.2  & \underline{54.7} & 79.7  & 80.8  & \underline{54.9} & 30.7  & \underline{44.6}  & 73.61  & 0.24 \\
    \midrule
    GRFT(ours)  & 69.5  & \textbf{93.6}  & \textbf{75.9} & \textbf{99.5}  & 91.4  & \textbf{91.2} & 52.2  & \textbf{88.2} & 96.0  & \underline{86.5}  & \textbf{76.3} & 81.4  & 62.3  & \textbf{55.1}  & 80.9  & \textbf{81.9} & \textbf{55.8}  & 32.0  & 43.6  & \textbf{74.38}  & 0.30 \\
    \bottomrule
    \end{tabular}}
  
  \label{Table:VTAB performances}
\end{table}%

% Table generated by Excel2LaTeX from sheet 'VTAB'

In Table \ref{Table:VTAB performances}, The GRFT achieves an impressive mean accuracy of 74.38\%, which is the highest among all the methods listed in the table. For instance, GRFT outperforms GPS*, which has a mean accuracy of 73.89\%, and SPT-LoRA, which has a mean accuracy of 74.07\%. The GRFT operates with a mean parameter percentage of 0.30\%, which is relatively low compared to some other methods. This suggests that GRFT can achieve high accuracy, making it more efficient in terms of computational resources. In summary, the GRFT stands out for its high accuracy and parameter efficiency across a wide range of datasets. Its ability to achieve superior results with fewer parameters makes it a promising candidate for applications where computational efficiency and model compactness are critical.

% We conducted tests on LLaMA3, and the results are shown in the Table \ref{Table:GLUE performances}. Due to the large storage requirements of GPS, it could not be run on LLaMA3. Therefore, we compared our method with full fine-tuning and LoRA. This fully demonstrates that our method exhibits strong adaptability across different models, ensuring generalization and enabling fine-tuning for models with various architectures.

We conducted experiments on LLaMA3, with the results presented in Table \ref{Table:GLUE performances}. Due to the high storage requirements of GPS, it was not feasible to run it on LLaMA3. Therefore, we compared our method with full fine-tuning and LoRA. The results clearly demonstrate that our approach exhibits strong adaptability across different models, ensuring generalization and enabling efficient fine-tuning across various model architectures. Additionally, our method significantly reduces computational and storage overhead while maintaining high performance, making it more practical for real-world applications with resource constraints.
\begin{table}[t]
  \centering
    \renewcommand{\arraystretch}{1.5}
      \caption{Comparisons results on sub-tasks of the GLUE with LLaMA3-1B models on a single GPU.}
    \resizebox{0.6\textwidth}{!}{
    \begin{tabular}{l|ccc|cc}
    \toprule
    \multicolumn{1}{c|}{Dataset} & \multicolumn{1}{c}{CoLA} & \multicolumn{1}{c}{MRPC} & \multicolumn{1}{c|}{RTE} & \multicolumn{1}{c}{Mean Acc.} & \multicolumn{1}{c}{Params.(\%)}\\
    \midrule
    Full  & 0.8428 & 0.8603 & 0.8087 & 0.8373 & 100.00 \\
    LoRA \cite{hu2021lora}  & 0.8562 & 0.8554 & 0.8159 & 0.8425 & 0.19 \\
    GRFT(ours) & 0.8495 & 0.8554 & 0.8484 & \textbf{0.8511} & \textbf{0.08} \\
    \bottomrule
    \end{tabular}}%

  \label{Table:GLUE performances}%
\end{table}%

\subsection{Ablation Study}

\begin{figure}[t]
	\centering
	\begin{minipage}{0.4\linewidth}
		\vspace{3pt}
        %这个图片路径替换成你的图片路径即可使用
		\centerline{\includegraphics[width=\textwidth]{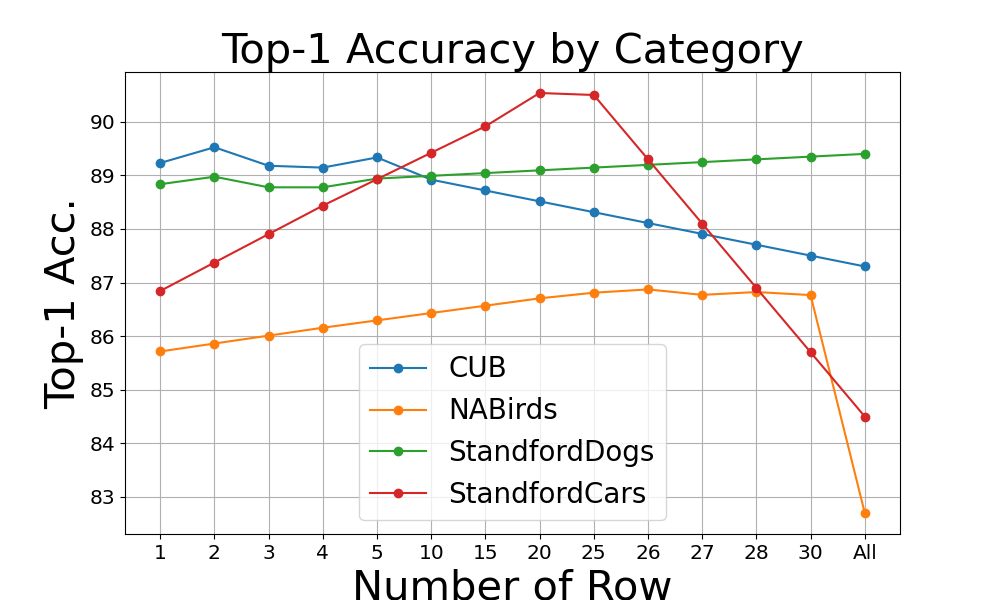}}
          % 加入对这列的图片说明
		% \centerline{Number of row}
	\end{minipage}
	\begin{minipage}{0.4\linewidth}
		\vspace{3pt}
		\centerline{\includegraphics[width=\textwidth]{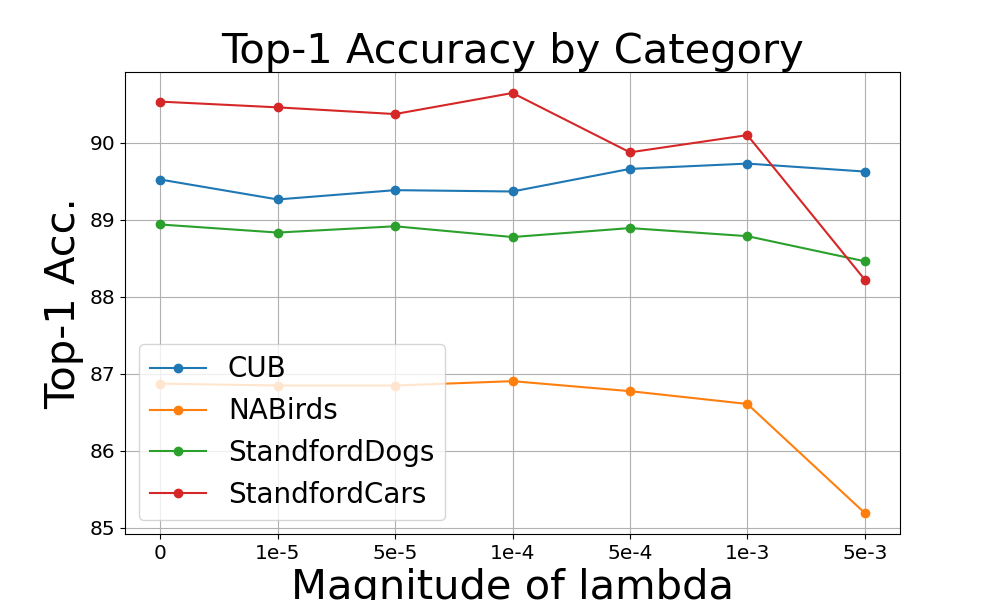}} 
		% \centerline{Magnitude of $\lambda$}
	\end{minipage}
	\begin{minipage}{0.4\linewidth}
		\vspace{3pt}
		\centerline{\includegraphics[width=\textwidth]{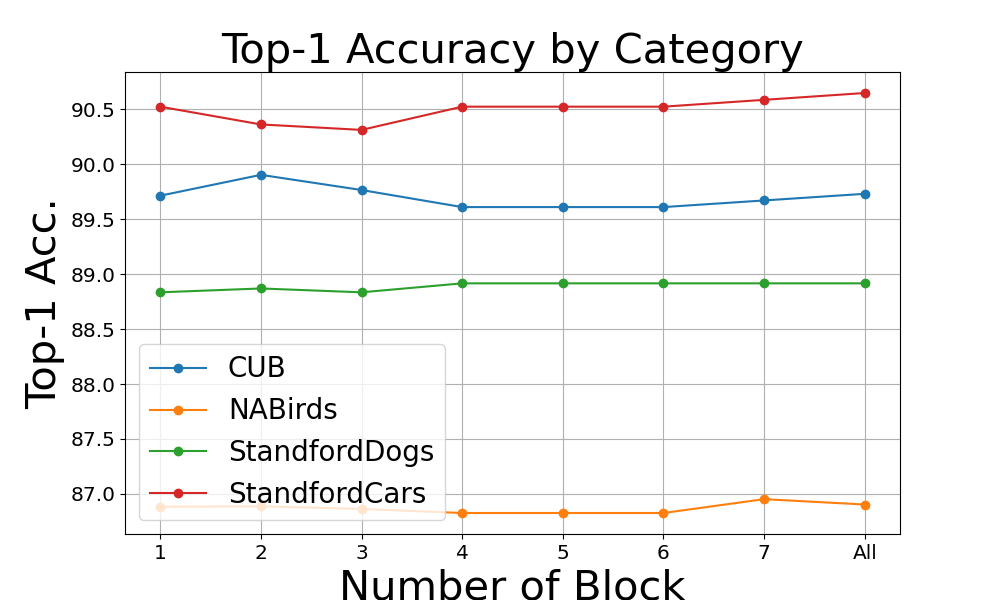}}	 
		% \centerline{Number of Block}
	\end{minipage}
        \begin{minipage}{0.4\linewidth}
		\vspace{3pt}
        %这个图片路径替换成你的图片路径即可使用
		\centerline{\includegraphics[width=\textwidth]{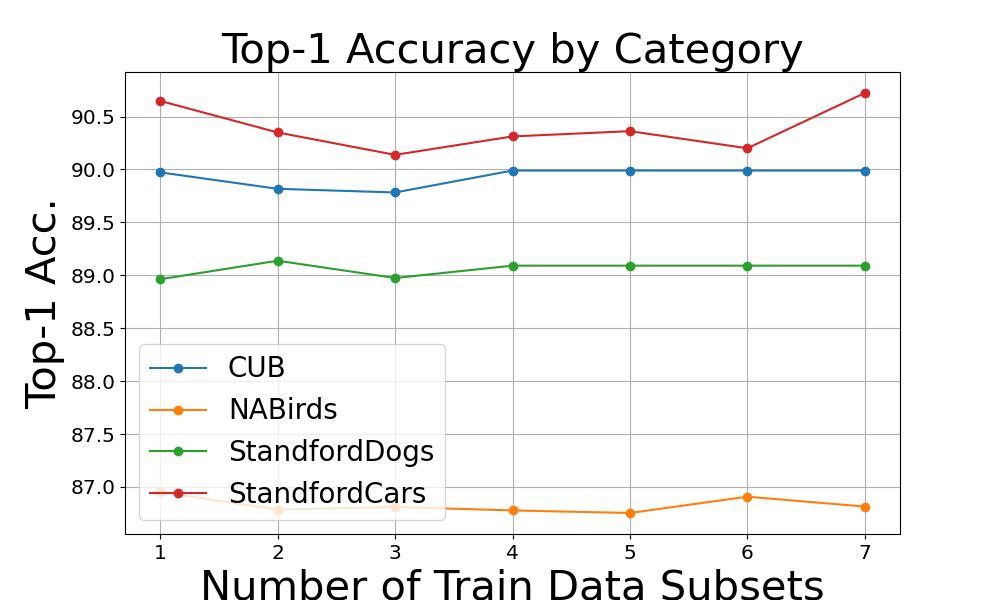}}
          % 加入对这列的图片说明
		% \centerline{Number of row}
	\end{minipage}
 
	\caption{Impacts of different setting. From left to right: (a) Impacts of different numbers of selected rows on performance. (b) Impacts of different Magnitude of lambda in $L_2$ norm on performance. (c) Impacts of different numbers of selected block on performance. (d) Impacts of different numbers of train data subsets on performance.}
	\label{Fig.ablation}
\end{figure}

\subsubsection{Parameters Selection}

\textbf{Selected Rows or Columns number}
We select the top $k$ rows or columns of each gradient matrix as the trainable parameters, with $k$ ranging from 1 to 30, and conduct experiments across multiple tasks. It can be observed that having more trainable parameters does not necessarily lead to better performance; instead, each dataset exhibits a performance peak. Furthermore, on larger datasets, adding trainable parameters significantly enhances accuracy. By controlling the number of trainable parameters, it is possible to achieve optimal results across different datasets. The Rows number results are shown in Fig. \ref{Fig.ablation} (a).

\textbf{Selected Methods: Sparse, Row, or Column?}
Our approach incorporates two selection methods: one based on selecting rows and the other on selecting columns. To investigate the differences among these two methods and sparse selection, we conducted an ablation experiment on the FGVC dataset, and the performance results are presented in Table \ref{tab:ablation of the different selected methods}. Our findings reveal that while both rows and columns selection types exhibit comparable overall performance, they yield different results depending on the specific characteristics of the data. And the row/column selection outperforms the sparse selection scheme. This suggests that the choice of rows/columns selection method may have varying impacts on model performance, influenced by the structure and nature of the dataset.

% To analyze the differences between row selection and column selection, we have conducted a simple preliminary analysis, as illustrated in the Fig. \ref{fig:row and column}. For a weight matrix $W$ and its corresponding gradient matrix of the same dimensions, selecting a row corresponds to the parameters that represent all input connections for a specific feature component in the output vector $y$, while selecting a column corresponds to all output connections for a specific feature component in the input vector $x$. We assume that row selection focuses on the complete feature from the previous layer to a specific feature component in the current layer, and during back propagation, it represents the impact of the feature component in the current layer on the previous layer. In contrast, column selection focuses on the influence from a specific feature component in the previous layer to the features in the next layer. The choice of which one to select in practice, along with the related principles, can be left for future research.
% \begin{figure}[h] 
%     \centering
%     \includegraphics[width=0.8\textwidth]{Figures/column or row 2.drawio.png} 
%     \caption{The different selection between Row and Column in Net. $x$ is the input feature, $h$ is the hidden layer feature, $y$ is the output feature and $\phi$ is the nonlinear function. The selected parameters in Row or Column are represented in blue.} 
%     \label{fig:row and column} 
% \end{figure}

% Table generated by Excel2LaTeX from sheet 'Sheet4'
\begin{table}[ht]
  \centering
  \renewcommand{\arraystretch}{1.5}
  \caption{Comparisons results on FGVC in different selected methods.}
  \resizebox{0.8\textwidth}{!}{
    \begin{tabular}{l|cccccc}
    \toprule
    \multicolumn{1}{c|}{Dataset} &
    \multicolumn{1}{c}{\centering CUB-2011} &
    \multicolumn{1}{c}{\centering NABirds} &
    \multicolumn{1}{c}{\centering Oxford Flowers} &
    \multicolumn{1}{c}{\centering Stan.Dogs} &
    \multicolumn{1}{c}{\centering Stan.Cars} & 
    \multicolumn{1}{c}{\centering Mean Acc.}  
    \\
    \midrule
    GPS* \cite{zhang2024gradient} & 89.6  & 86.8  & 99.7  & 88.9  & 90.4  & 91.06 \\
    \midrule
    GRFT-Sparse & 89.7  & 86.8  & 99.7  & 88.9  & 90.6  & 91.13  \\
    GRFT-Row & 90.0  & 87.0  & 99.7  & 89.1  & 90.7  & 91.29\\
    GRFT-Column & 90.1  & 86.9  & 99.7  & 88.8  & 90.8  & 91.27\\
    \bottomrule
    \end{tabular}}%
  \label{tab:ablation of the different selected methods}%
\end{table}%

\subsubsection{Regularization}

\textbf{Types of Regularization Norms}
The aims of regularization are two aspects. Firstly, it ensures that the model parameters are updated in the vicinity of the pre-trained model's weights, facilitating the transfer of knowledge from the pre-trained model. Secondly, it enhances the model's generalization ability. There are various choices of regularization norms, and here we focus on comparing the performance differences between $L_1$ and $L_2$ norms under the same parameter settings. The experimental results are shown in Table \ref{tab:ablation of the different norms}. From the results, we observe that $L_2$ regularization generally outperforms $L_1$ regularization.
% Table generated by Excel2LaTeX from sheet 'Sheet4'
\begin{table}[ht]
  \centering
  \renewcommand{\arraystretch}{1.5}
  \caption{Comparisons results on FGVC in different norms.}
  \resizebox{0.6\textwidth}{!}{
    \begin{tabular}{l|cccccc}
    \toprule
    \multicolumn{1}{c|}{Dataset} &
    \multicolumn{1}{c}{\centering CUB-2011} &
    \multicolumn{1}{c}{\centering NABirds} &
    \multicolumn{1}{c}{\centering Oxford Flowers} &
    \multicolumn{1}{c}{\centering Stan.Dogs} &
    \multicolumn{1}{c}{\centering Stan.Cars} & 
    \multicolumn{1}{c}{\centering Mean Acc.} 
    \\
    \midrule
    Without Norm & 89.5  & 86.9  & 99.6  & 88.9  & 90.5  & 91.10 \\
    \midrule
    $L_1$ Norm & 89.8  & 85.8  & 99.6  & \multicolumn{1}{c}{88.8 } & 88.7  & 90.54\\
    $L_2$ Norm & 89.9  & 86.9  & 99.7  & 89.1  & 90.6  & 91.25\\
    \bottomrule
    \end{tabular}}%
  \label{tab:ablation of the different norms}%
\end{table}%

\textbf{Magnitude of Regularization Parameter}
After determining the specific selection and regularization methods, we conducted parameter tests for regularization. We tested the magnitude of $\lambda$, selecting values within the range of $1 \times 10^{-8}$ to $1 \times 10^{-3}$, and performed experiments for each task. The results Fig.\ref{Fig.ablation} (b) indicate that each dataset has an optimal $\lambda$ that yields the best performance. 

\textbf{Number of Regular Blocks} 
Once $\lambda$ was determined, we further tested the number of regular blocks, ranging from 1 to 8, and selected the best performing configuration as the optimal result for the dataset under our proposed approach Fig.\ref{Fig.ablation} (c).

\subsubsection{Data Processing}
In our approach, we perform a simple processing step on the datasets. Specifically, we randomly divide the training dataset into smaller subsets and choose the subset with the minimal loss to calculate the mask and determine the parameters to be trained or kept frozen. Thurs, there exits a hyperparameter $n$, which represents the number of subsets into which the dataset is divided. We evaluate the performance under different values of this parameter, ranging from 1 to 7, where 1 indicates that no partitioning is applied. The experimental results are presented in Fig.\ref{Fig.ablation} (d).

\section{Conclusion}
In this paper, we proposed gradient-based and regularized fine-tuning (GRFT), an innovative parameter-efficient fine-tuning method. GRFT selectively updates the parameters corresponding to rows or columns with the largest sum of squared gradients, while incorporating $L_2$ regularization to mitigate the challenges of computational and storage inefficiency and to preserve knowledge when adapting large pre-trained models to downstream tasks. GRFT demonstrates significant improvements in average accuracy on FGVC and VTAB, outperforming existing parameter-efficient fine-tuning (PEFT) methods. Future work can explore the integration of GRFT with continual learning techniques to enable lifelong adaptation across evolving tasks without excessive computational cost.

% \newpage
% \today
{\small
 \bibliographystyle{ieee}
 \bibliography{egbib}

\begin{thebibliography}{10}\itemsep=-1pt

\bibitem{Cheng_2017}
G.~Cheng, J.~Han, and X.~Lu.
\newblock Remote sensing image scene classification: Benchmark and state of the art.
\newblock {\em Proceedings of the IEEE}, 105(10):1865–1883, Oct. 2017.

\bibitem{Deng_Dong_Socher_Li_Kai_-Li_Li_Fei-Fei_2009}
J.~Deng, W.~Dong, R.~Socher, L.-J. Li, K.~Li, and L.~Fei-Fei.
\newblock Imagenet: A large-scale hierarchical image database.
\newblock In {\em 2009 IEEE Conference on Computer Vision and Pattern Recognition}, Jun 2009.

\bibitem{Dosovitskiy_Beyer_Kolesnikov_Weissenborn_Zhai_Unterthiner_Dehghani_Minderer_Heigold_Gelly_etal._2020}
A.~Dosovitskiy, L.~Beyer, A.~Kolesnikov, D.~Weissenborn, X.~Zhai, T.~Unterthiner, M.~Dehghani, M.~Minderer, G.~Heigold, S.~Gelly, J.~Uszkoreit, and N.~Houlsby.
\newblock An image is worth 16x16 words: Transformers for image recognition at scale.
\newblock {\em arXiv: Computer Vision and Pattern Recognition,arXiv: Computer Vision and Pattern Recognition}, Oct 2020.

\bibitem{1384978}
L.~Fei-Fei, R.~Fergus, and P.~Perona.
\newblock Learning generative visual models from few training examples: An incremental bayesian approach tested on 101 object categories.
\newblock In {\em 2004 Conference on Computer Vision and Pattern Recognition Workshop}, pages 178--178, 2004.

\bibitem{Frankle_Carbin_2018}
J.~Frankle and M.~Carbin.
\newblock The lottery ticket hypothesis: Finding sparse, trainable neural networks.
\newblock {\em arXiv: Learning,arXiv: Learning}, Mar 2018.

\bibitem{FRENCH1999128}
R.~M. French.
\newblock Catastrophic forgetting in connectionist networks.
\newblock {\em Trends in Cognitive Sciences}, 3(4):128--135, 1999.

\bibitem{Gale_Elsen_Hooker_2019}
T.~Gale, E.~Elsen, and S.~Hooker.
\newblock The state of sparsity in deep neural networks.
\newblock {\em arXiv: Learning,arXiv: Learning}, Feb 2019.

\bibitem{grattafiori2024llama3herdmodels}
A.~Grattafiori and A.~D. etc.
\newblock The llama 3 herd of models, 2024.

\bibitem{Han_Mao_Dally_2015}
S.~Han, H.~Mao, and W.~Dally.
\newblock Deep compression: Compressing deep neural networks with pruning, trained quantization and huffman coding.
\newblock {\em arXiv: Computer Vision and Pattern Recognition,arXiv: Computer Vision and Pattern Recognition}, Oct 2015.

\bibitem{Han_Pool_Tran_Dally_2015}
S.~Han, J.~Pool, J.~Tran, and W.~Dally.
\newblock Learning both weights and connections for efficient neural networks.
\newblock {\em arXiv: Neural and Evolutionary Computing,arXiv: Neural and Evolutionary Computing}, Jun 2015.

\bibitem{hayou2024lora+}
S.~Hayou, N.~Ghosh, and B.~Yu.
\newblock Lora+: Efficient low rank adaptation of large models.
\newblock {\em arXiv preprint arXiv:2402.12354}, 2024.

\bibitem{he2023sensitivityawarevisualparameterefficientfinetuning}
H.~He, J.~Cai, J.~Zhang, D.~Tao, and B.~Zhuang.
\newblock Sensitivity-aware visual parameter-efficient fine-tuning, 2023.

\bibitem{he2015delving}
K.~He, X.~Zhang, S.~Ren, and J.~Sun.
\newblock Delving deep into rectifiers: Surpassing human-level performance on imagenet classification.
\newblock In {\em Proceedings of the IEEE international conference on computer vision}, pages 1026--1034, 2015.

\bibitem{Hoerl2000RidgeRB}
A.~E. Hoerl and R.~W. Kennard.
\newblock Ridge regression: Biased estimation for nonorthogonal problems.
\newblock {\em Technometrics}, 42:80 -- 86, 2000.

\bibitem{houlsby2019parameterefficienttransferlearningnlp}
N.~Houlsby, A.~Giurgiu, S.~Jastrzebski, B.~Morrone, Q.~de~Laroussilhe, A.~Gesmundo, M.~Attariyan, and S.~Gelly.
\newblock Parameter-efficient transfer learning for nlp, 2019.

\bibitem{Houlsby_Giurgiu_Jastrzębski_Morrone_Laroussilhe_Gesmundo_Attariyan_Gelly_2019}
N.~Houlsby, A.~Giurgiu, S.~Jastrzębski, B.~Morrone, Q.~Laroussilhe, A.~Gesmundo, M.~Attariyan, and S.~Gelly.
\newblock Parameter-efficient transfer learning for nlp.
\newblock {\em International Conference on Machine Learning,International Conference on Machine Learning}, Jan 2019.

\bibitem{hu2021lora}
E.~J. Hu, Y.~Shen, P.~Wallis, Z.~Allen-Zhu, Y.~Li, S.~Wang, L.~Wang, and W.~Chen.
\newblock Lora: Low-rank adaptation of large language models.
\newblock {\em arXiv preprint arXiv:2106.09685}, 2021.

\bibitem{jia2022visual}
M.~Jia, L.~Tang, B.-C. Chen, C.~Cardie, S.~Belongie, B.~Hariharan, and S.-N. Lim.
\newblock Visual prompt tuning.
\newblock In {\em European Conference on Computer Vision}, pages 709--727. Springer, 2022.

\bibitem{johnson2017clevr}
J.~Johnson, B.~Hariharan, L.~van~der Maaten, L.~Fei-Fei, C.~L. Zitnick, and R.~Girshick.
\newblock Clevr: A diagnostic dataset for compositional language and elementary visual reasoning.
\newblock In {\em CVPR}, 2017.

\bibitem{karimi2021compacter}
R.~Karimi~Mahabadi, J.~Henderson, and S.~Ruder.
\newblock Compacter: Efficient low-rank hypercomplex adapter layers.
\newblock {\em Advances in Neural Information Processing Systems}, 34:1022--1035, 2021.

\bibitem{KhoslaYaoJayadevaprakashFeiFei_FGVC2011}
A.~Khosla, N.~Jayadevaprakash, B.~Yao, and L.~Fei-Fei.
\newblock Novel dataset for fine-grained image categorization.
\newblock In {\em First Workshop on Fine-Grained Visual Categorization, IEEE Conference on Computer Vision and Pattern Recognition}, Colorado Springs, CO, June 2011.

\bibitem{Khosla_Teterwak_Wang_Sarna_Tian_Isola_Maschinot_Liu_Krishnan_2020}
P.~Khosla, P.~Teterwak, C.~Wang, A.~Sarna, Y.~Tian, P.~Isola, A.~Maschinot, C.~Liu, and D.~Krishnan.
\newblock Supervised contrastive learning.
\newblock {\em Neural Information Processing Systems,Neural Information Processing Systems}, Apr 2020.

\bibitem{Kingma_Ba_2014}
D.~Kingma and J.~Ba.
\newblock Adam: A method for stochastic optimization.
\newblock {\em arXiv: Learning,arXiv: Learning}, Dec 2014.

\bibitem{6755945}
J.~Krause, M.~Stark, J.~Deng, and L.~Fei-Fei.
\newblock 3d object representations for fine-grained categorization.
\newblock In {\em 2013 IEEE International Conference on Computer Vision Workshops}, pages 554--561, 2013.

\bibitem{krizhevsky2009learning}
A.~Krizhevsky, G.~Hinton, et~al.
\newblock Learning multiple layers of features from tiny images.
\newblock 2009.

\bibitem{Kruschke_Movellan_1991}
J.~Kruschke and J.~Movellan.
\newblock Benefits of gain: speeded learning and minimal hidden layers in back-propagation networks.
\newblock {\em IEEE Transactions on Systems, Man, and Cybernetics}, 21(1):273–280, Jan 1991.

\bibitem{kurtic2023ziplminferenceawarestructuredpruning}
E.~Kurtic, E.~Frantar, and D.~Alistarh.
\newblock Ziplm: Inference-aware structured pruning of language models, 2023.

\bibitem{Li_Kadav_Durdanovic_Samet_Graf_2016}
H.~Li, A.~Kadav, I.~Durdanovic, H.~Samet, and H.~Graf.
\newblock Pruning filters for efficient convnets.
\newblock {\em arXiv: Computer Vision and Pattern Recognition,arXiv: Computer Vision and Pattern Recognition}, Aug 2016.

\bibitem{li2024large}
J.~Li, J.~Xu, S.~Huang, Y.~Chen, W.~Li, J.~Liu, Y.~Lian, J.~Pan, L.~Ding, H.~Zhou, et~al.
\newblock Large language model inference acceleration: A comprehensive hardware perspective.
\newblock {\em arXiv preprint arXiv:2410.04466}, 2024.

\bibitem{10688138}
Q.~Li.
\newblock Parameter efficient fine-tuning on selective parameters for transformer-based pre-trained models.
\newblock In {\em 2024 IEEE International Conference on Multimedia and Expo (ICME)}, pages 1--6, 2024.

\bibitem{lian2023scalingshiftingfeatures}
D.~Lian, D.~Zhou, J.~Feng, and X.~Wang.
\newblock Scaling \& shifting your features: A new baseline for efficient model tuning, 2023.

\bibitem{loshchilov2019decoupledweightdecayregularization}
I.~Loshchilov and F.~Hutter.
\newblock Decoupled weight decay regularization, 2019.

\bibitem{lu2023uniadapter}
H.~Lu, M.~Ding, Y.~Huo, G.~Yang, Z.~Lu, M.~Tomizuka, and W.~Zhan.
\newblock Uniadapter: Unified parameter-efficient transfer learning for cross-modal modeling.
\newblock {\em arXiv preprint arXiv:2302.06605}, 2023.

\bibitem{Nilsback08}
M.-E. Nilsback and A.~Zisserman.
\newblock Automated flower classification over a large number of classes.
\newblock In {\em Indian Conference on Computer Vision, Graphics and Image Processing}, Dec 2008.

\bibitem{openai2024gpt4technicalreport}
OpenAI, J.~Achiam, S.~Adler, S.~Agarwal, L.~Ahmad, I.~Akkaya, and F.~L. Aleman.etc.
\newblock Gpt-4 technical report, 2024.

\bibitem{Ouyang_Wang_Zhang_Yang_2016}
W.~Ouyang, X.~Wang, C.~Zhang, and X.~Yang.
\newblock Factors in finetuning deep model for object detection with long-tail distribution.
\newblock In {\em 2016 IEEE Conference on Computer Vision and Pattern Recognition (CVPR)}, Jun 2016.

\bibitem{qiang2024bilora}
R.~Qiang, R.~Zhang, and P.~Xie.
\newblock Bilora: A bi-level optimization framework for overfitting-resilient low-rank adaptation of large pre-trained models.
\newblock {\em arXiv preprint arXiv:2403.13037}, 2024.

\bibitem{toneva2018empirical}
M.~Toneva, A.~Sordoni, R.~T.~d. Combes, A.~Trischler, Y.~Bengio, and G.~J. Gordon.
\newblock An empirical study of example forgetting during deep neural network learning.
\newblock {\em arXiv preprint arXiv:1812.05159}, 2018.

\bibitem{touvron2023llama2openfoundation}
H.~Touvron, L.~Martin, K.~Stone, P.~Albert, A.~Almahairi, Y.~Babaei, N.~Bashlykov, S.~Batra, P.~Bhargava, and S.~Bhosale.etc.
\newblock Llama 2: Open foundation and fine-tuned chat models, 2023.

\bibitem{Horn_2015_CVPR}
G.~Van~Horn, S.~Branson, R.~Farrell, S.~Haber, J.~Barry, P.~Ipeirotis, P.~Perona, and S.~Belongie.
\newblock Building a bird recognition app and large scale dataset with citizen scientists: The fine print in fine-grained dataset collection.
\newblock In {\em Proceedings of the IEEE Conference on Computer Vision and Pattern Recognition (CVPR)}, June 2015.

\bibitem{Veeling2018-qh}
B.~S. Veeling, J.~Linmans, J.~Winkens, T.~Cohen, and M.~Welling.
\newblock Rotation equivariant {CNNs} for digital pathology.
\newblock June 2018.

\bibitem{Wah2011TheCB}
C.~Wah, S.~Branson, P.~Welinder, P.~Perona, and S.~J. Belongie.
\newblock The caltech-ucsd birds-200-2011 dataset.
\newblock 2011.

\bibitem{wang2019gluemultitaskbenchmarkanalysis}
A.~Wang, A.~Singh, J.~Michael, F.~Hill, O.~Levy, and S.~R. Bowman.
\newblock Glue: A multi-task benchmark and analysis platform for natural language understanding, 2019.

\bibitem{wang2023lion}
H.~Wang, J.~Chang, X.~Luo, J.~Sun, Z.~Lin, and Q.~Tian.
\newblock Lion: Implicit vision prompt tuning.
\newblock {\em arXiv preprint arXiv:2303.09992}, 2023.

\bibitem{Wen_Wu_Wang_Chen_Li_2016}
W.~Wen, C.~Wu, Y.~Wang, Y.~Chen, and H.~Li.
\newblock Learning structured sparsity in deep neural networks.
\newblock {\em Neural Information Processing Systems,Neural Information Processing Systems}, Aug 2016.

\bibitem{zaken2022bitfitsimpleparameterefficientfinetuning}
E.~B. Zaken, S.~Ravfogel, and Y.~Goldberg.
\newblock Bitfit: Simple parameter-efficient fine-tuning for transformer-based masked language-models, 2022.

\bibitem{zhai2020largescalestudyrepresentationlearning}
X.~Zhai, J.~Puigcerver, A.~Kolesnikov, P.~Ruyssen, C.~Riquelme, M.~Lucic, J.~Djolonga, A.~S. Pinto, M.~Neumann, A.~Dosovitskiy, L.~Beyer, O.~Bachem, M.~Tschannen, M.~Michalski, O.~Bousquet, S.~Gelly, and N.~Houlsby.
\newblock A large-scale study of representation learning with the visual task adaptation benchmark, 2020.

\bibitem{zhang2023llama}
R.~Zhang, J.~Han, C.~Liu, P.~Gao, A.~Zhou, X.~Hu, S.~Yan, P.~Lu, H.~Li, and Y.~Qiao.
\newblock Llama-adapter: Efficient fine-tuning of language models with zero-init attention.
\newblock {\em arXiv preprint arXiv:2303.16199}, 2023.

\bibitem{zhang2024gradient}
Z.~Zhang, Q.~Zhang, Z.~Gao, R.~Zhang, E.~Shutova, S.~Zhou, and S.~Zhang.
\newblock Gradient-based parameter selection for efficient fine-tuning.
\newblock In {\em Proceedings of the IEEE/CVF Conference on Computer Vision and Pattern Recognition}, pages 28566--28577, 2024.

\bibitem{zhou2022learning}
K.~Zhou, J.~Yang, C.~C. Loy, and Z.~Liu.
\newblock Learning to prompt for vision-language models.
\newblock {\em International Journal of Computer Vision}, 130(9):2337--2348, 2022.

\end{thebibliography}
}

% \newpage
\appendix
% \onecolumn
\section{Appendix}
\subsection{Baseline Description}
\subsubsection{GPS}
GPS (Gradient-based Parameter Selection) is an innovative Parameter-Efficient Fine-Tuning (PEFT) method designed to address the computational and storage challenges associated with fine-tuning large-scale pretrained models on downstream tasks. Compared to traditional full-parameter fine-tuning approaches, GPS achieves efficient model adaptation by adjusting only a small subset of key parameters in the pretrained model while keeping the remaining parameters frozen. This significantly reduces computational costs and memory consumption.

The core idea of GPS is to select the most critical parameters for a downstream task based on gradient information. Specifically, the method first computes the gradient values of each neuron’s input connections, where the magnitude of the gradient reflects the importance of the parameter in the current task. GPS selects parameters with the highest gradient values, as these parameters exhibit the most rapid changes in the loss function and contribute the most to model performance improvement. Additionally, to ensure that the model can adjust to features at different levels, GPS employs a distributed parameter selection strategy—rather than simply selecting the parameters with the highest gradients across the entire network, it selects the top input connections within each neuron. This strategy ensures a more balanced parameter distribution across different layers of the model, allowing for better adaptation to the feature requirements of downstream tasks.

GPS offers several significant advantages. First, it does not introduce any additional parameters, thereby avoiding increased computational overhead during both training and inference. Second, GPS is model-agnostic and can be applied to various architectures, such as Transformers and CNNs, without requiring modifications to the model structure. Furthermore, GPS dynamically selects parameters based on the specific needs of each downstream task, leading to improved adaptability and overall performance.

\subsubsection{LoRA}
LoRA (Low-Rank Adaptation) is an efficient parameter adaptation method specifically designed for fine-tuning large-scale pre-trained language models. It adapts to downstream tasks by injecting trainable low-rank factorized matrices into each layer of the Transformer architecture, while keeping the pre-trained weights frozen. This significantly reduces the number of trainable parameters required for the downstream task. The core idea of LoRA is based on the assumption that the weight changes during model adaptation have low "intrinsic rank," meaning they can be approximated by low-rank matrices. This approach allows LoRA to substantially reduce computational and storage costs while maintaining model performance, and it does not introduce additional inference latency during deployment. LoRA has shown outstanding performance across multiple natural language processing tasks, being competitive with full-parameter fine-tuning in terms of model quality, while significantly reducing the number of trainable parameters and GPU memory requirements.

\subsection{The Differences Between GPS And GRFT}

\subsubsection{Parameter Selection Strategy}

\begin{figure}[htbp] 
    \centering
    \includegraphics[width=1\textwidth]{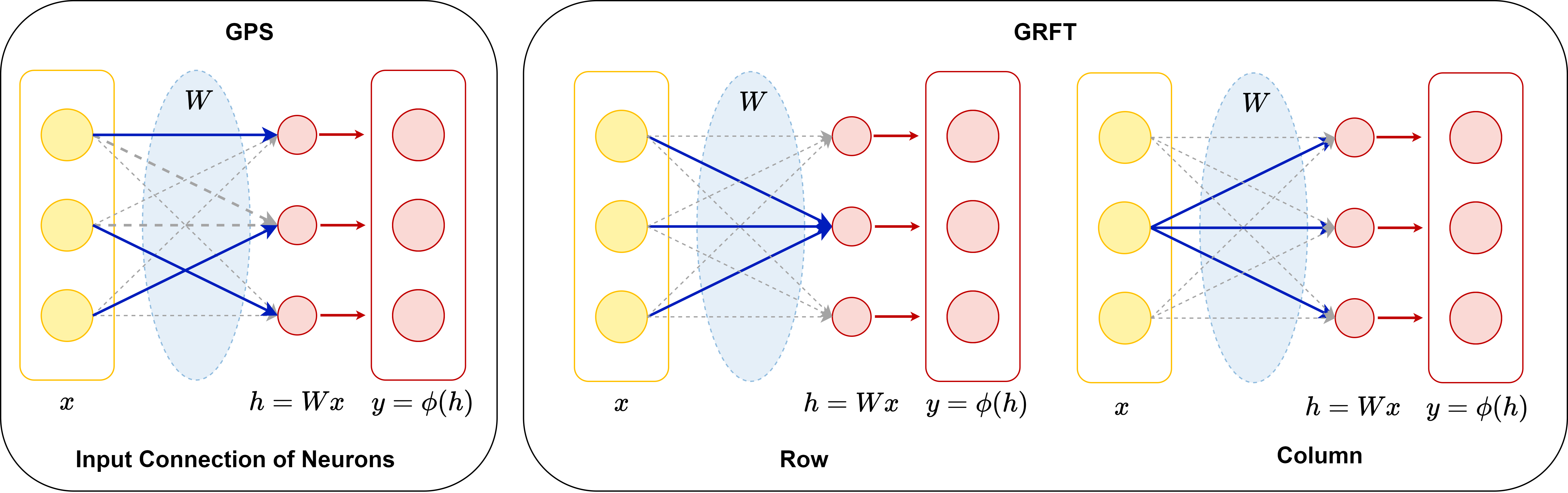} 
    \caption{The differences selection methods between GPS and GRFT. The illustration depicts the parameter selection when $K=1$ for both methods, which marked in blue. Specifically, GPS selects one connection per neuron, while GRFT selects an entire row or column. The selection pattern of GPS is sparse, meaning that each row of the parameter matrix updates only a single element. In contrast, GRFT updates an entire row of the weight matrix at once.} 
    \label{fig:difference between GPS and GRFT} 
\end{figure}

The differences of strategy is shown as Fig.\ref{fig:difference between GPS and GRFT}.

\begin{itemize}
    \item \textbf{GPS:} GPS selects parameters for fine-tuning by computing the gradient values of each neuron's input connections. For each neuron, it selects the top-$K$ parameters with the highest gradient values. This approach ensures that the selected parameters undergo the most rapid changes in the loss function, enabling the model to quickly adapt during fine-tuning. GPS relies on a sparse matrix-based parameter selection strategy, requiring a mask matrix of the same dimensions as the weight matrix, which increases storage costs.
    
    \item \textbf{GRFT:} GRFT selects entire rows or columns of the weight matrix for fine-tuning instead of individual parameters. Specifically, it selects the top-$K$ rows or columns with the highest gradient squared sum. This method reduces storage costs as it only requires storing the indices of the selected rows or columns, rather than a full mask matrix. GRFT’s parameter selection strategy is more efficient in hardware implementation since it avoids the computational complexity associated with sparse matrix operations.
\end{itemize}

To analyze the differences between row selection and column selection, we have conducted a simple preliminary analysis, as illustrated in the Fig. \ref{fig:difference between GPS and GRFT}. For a weight matrix $W$ and its corresponding gradient matrix of the same dimensions, selecting a row corresponds to the parameters that represent all input connections for a specific feature component in the output vector $y$, while selecting a column corresponds to all output connections for a specific feature component in the input vector $x$. We assume that row selection focuses on the complete feature from the previous layer to a specific feature component in the current layer, and during back propagation, it represents the impact of the feature component in the current layer on the previous layer. In contrast, column selection focuses on the influence from a specific feature component in the previous layer to the features in the next layer. The choice of which one to select in practice, along with the related principles, can be left for future research.

\subsubsection{Regularization Strategy}
\begin{itemize}
    \item \textbf{GPS:} GPS does not introduce additional regularization strategies. It primarily relies on gradient selection for parameter optimization.

    \item \textbf{GRFT:} GRFT incorporates $L_2$ regularization by adding a regularization term to the loss function. This constrains the updates of fine-tuned parameters to remain close to those of the pre-trained model. The regularization strategy helps prevent excessive parameter adjustments during fine-tuning, preserves the knowledge acquired in the pre-training phase, and mitigates catastrophic forgetting.
\end{itemize}

\subsubsection{Storage}
\begin{itemize}
    \item \textbf{GPS:} GPS requires storing a mask matrix of the same dimensions as the weight matrix, leading to increased storage costs.

    \item \textbf{GRFT:} GRFT only requires storing the indices of selected rows or columns, significantly reducing storage costs.
\end{itemize}

\subsection{Experiment Details}
% Table generated by Excel2LaTeX from sheet 'Experiment Detial'
\begin{table}[htbp]
  \centering
  \renewcommand{\arraystretch}{1.5}
      \caption{Hyperparameters on FGVC}
  \resizebox{0.8\textwidth}{!}{
    \begin{tabular}{l|ccccccc}
    \toprule
    \multicolumn{1}{c|}{Dataset} & \multicolumn{1}{m{4.055em}}{\centering Learning\newline{}Rate} & \multicolumn{1}{m{2.78em}}{\centering Batch\newline{}size} & Epoch & \multicolumn{1}{m{6em}}{\centering Data Subsets\newline{}Number} & \multicolumn{1}{m{5.445em}}{\centering Regular\newline{}Parameter} & \multicolumn{1}{m{6.28em}}{\centering Regular Layer\newline{}Number} & \multicolumn{1}{m{6.055em}}{\centering Row/Column\newline{}Number} \\
    \midrule
    CUB-2011 & 5e-3 & 32    & 100   & 4     & 1e-3 & 2     & 2\\
    NaBirds & 1e-4 & 32    & 100   & 1     & 1e-4 & 7     & 26 \\
    Oxford Flowers & 1e-3 & 32    & 100   & 1     & 1e-4 & 2     & 1 \\
    Stan. Dogs & 2e-4 & 64    & 100   & 2     & 5e-5 & 4     & 5 \\
    Stan. Cars & 5e-4 & 32    & 100   & 7     & 1e-4 & all   & 20\\
    \bottomrule
    \end{tabular}}%
  \label{tab:FGVC datail}%
\end{table}%

In this section, we present the relevant experimental parameter settings for our image classification tasks on FGVC and VTAB. The tables below include several parameters associated with the methods discussed in the paper. Their specific meanings are as follows: Data Subsets Number denotes the number of splits in the training dataset during data processing, Regular Parameter refers to the scale parameter in regularization, Regular Layer Number indicates the number of modules added for regularization constraints, with all presenting the all layers in model being added for regularization, and Row/Column Number represents the number of specific rows/columns selected during the fine-tuning process, corresponding to the number of training parameters.
\subsubsection{Experiments on FGVC}
We provide a detailed description of the experimental setup and results on the FGVC task. We list the hyperpamameters of the best performance on FGVC in Table \ref{tab:FGVC datail}.

\subsubsection{Experiments on VTAB}
We present explanations of the experimental setup and results for the VTAB task. We list the hyperpamameters of the best performance on VTAB in Table \ref{tab:VTAB datail}.
% Table generated by Excel2LaTeX from sheet 'Experiment Detial'
\begin{table}[htbp]
  \centering
  \renewcommand{\arraystretch}{1.5}
      \caption{Hyperparameters on VTAB}
  \resizebox{0.9\textwidth}{!}{
    \begin{tabular}{l|ccccccc}
    \toprule
    \multicolumn{1}{c|}{Dataset} & \multicolumn{1}{m{4.055em}}{\centering Learning\newline{}Rate} & \multicolumn{1}{m{2.78em}}{\centering Batch\newline{}size} & Epoch & \multicolumn{1}{m{6em}}{\centering Data Subsets\newline{}Number} & \multicolumn{1}{m{5.445em}}{\centering Regular\newline{}Parameter} & \multicolumn{1}{m{6.28em}}{\centering Regular Layer\newline{}Number} & \multicolumn{1}{m{6.055em}}{\centering Row/Column\newline{}Number} \\
    \midrule
    CIFAR-100 & 2e-3 & 32    & 100   & 1     & 1e-6 & all   & 1 \\
    Caltech101 & 2e-3 & 16    & 100   & 1     & 1e-8 & 2     & 1 \\
    DTD   & 2e-3 & 16    & 100   & 1     & 1e-8 & 3     & 1 \\
    Flowers102 & 2e-3 & 16    & 100   & 1     & 1e-6 & all   & 1 \\
    Pets  & 3e-3 & 32    & 100   & 1     & 1e-3 & all   & 2 \\
    SVHN  & 5e-3 & 32    & 100   & 1     & 1e-3 & 1     & 4 \\
    Sun397 & 2.5e-3 & 16    & 100   & 1     & 1e-6 & 2     & 1 \\
    Patch Camelyon & 4e-3 & 32    & 100   & 1     & 1e-7 & all   & 2 \\
    EuroSAT & 2e-3 & 16    & 100   & 1     & 1e-4 & 3     & 2 \\
    Resisc45 & 1.5e-3 & 16    & 100   & 1     & 1e-3 & all   & 3 \\
    Retinopathy & 2e-3 & 32    & 100   & 1     & 1e-6 & 1     & 1 \\
    Clevr/count & 3e-4 & 16    & 100   & 1     & 0     & 0     & 3 \\
    Clevr/distance & 2e-3 & 16    & 100   & 1     & 1e-5 & all   & 2 \\
    DMLab & 1.5e-3 & 16    & 100   & 1     & 0     & 0     & 2 \\
    KITTI/distance & 1e-3 & 16    & 100   & 1     & 1e-5 & all   & 1 \\
    dSprites/loc & 7e-3 & 32    & 100   & 1     & 1e-4 & 5     & 3 \\
    dSprites/ori & 5e-4 & 16    & 100   & 1     & 0     & 0     & 2 \\
    SmallNORB/azi & 3e-3 & 32    & 100   & 1     & 1e-8 & all   & 3 \\
    SmallNORB/ele & 3e-4 & 32    & 100   & 1     & 1e-3 & 2     & 2 \\
    \bottomrule
    \end{tabular}}%
  \label{tab:VTAB datail}%
\end{table}%

\subsubsection{Experiments on GLUE}
For GLUE, we test only three datasets for training to demonstrate the generalizability of our method across different models. Since we used the Llama 3 model and ran it on a single GPU, GPS requires storing a mask of the same size as the model, leading to excessive memory usage that caused the experiment to be unfeasible. Therefore, we compared our method with the full model and LoRA. This comparison further highlights the applicability of GRFT in large models.

\begin{table}[h]
  \centering
  \renewcommand{\arraystretch}{1.5}
      \caption{The number of learnable parameters across all tasks. }
  \resizebox{0.9\textwidth}{!}{
    \begin{tabular}{lcc|lcc|lcc}
    \toprule
    \multicolumn{1}{c}{Dataset} & \multicolumn{1}{m{3.72em}}{\centering Params.\newline{}(\%)} & \multicolumn{1}{m{6.055em}|}{\centering Row/Column\newline{}Number} & \multicolumn{1}{c}{Dataset} & \multicolumn{1}{m{4.78em}}{\centering Params.\newline{}(\%)} & \multicolumn{1}{m{6.055em}|}{\centering Row/Column\newline{}Number} & \multicolumn{1}{c}{Dataset} & \multicolumn{1}{m{3.72em}}{\centering Params.\newline{}(\%)} & \multicolumn{1}{m{6.055em}}{\centering Row/Column\newline{}Number} \\
    \midrule
    CUB-2011 & 0.47  & 2     & Pets  & 0.28  & 2     & DMLab & 0.30  & 2 \\
    NaBirds & 2.56  & 26    & SVHN  & 0.41  & 4     & KITTI/distance & 0.18  & 1 \\
    Oxford.Flowers & 0.26  & 1     & Sun397 & 0.53  & 1     & dSprites/loc & 0.34  & 3 \\
    Stan.Dogs & 0.58  & 5     & Patch Camelyon & 0.25  & 2     & dSprites/ori & 0.31  & 2 \\
    Stan.Cars & 2.22  & 20    & EuroSAT & 0.26  & 2     & SmallNORB/azi & 0.34  & 3 \\
    CIFAR-100 & 0.26  & 1     & Resisc45 & 0.37  & 3     & SmallNORB/ele & 0.30  & 2 \\
    Caltech101 & 0.26  & 1     & Retinopathy & 0.18  & 1     & CoLA  & 0.08  & 3 \\
    DTD   & 0.22  & 1     & Clevr/count & 0.33  & 3     & MRPC  & 0.08  & 3 \\
    Flowers102 & 0.26  & 1     & Clevr/distance & 0.26  & 2     & RTE   & 0.08  & 3 \\
    \bottomrule
    \end{tabular}}%
  \label{tab:The number of learnable parameters across all tasks}%
\end{table}%

\subsection{The Number of Training Parameters on Different Tasks}
For neural networks, our method selects entire rows or columns of parameters, as shown in the Fig.\ref{fig:difference between GPS and GRFT}. For datasets with larger volumes of data, such as Nabirds, we can choose more rows and columns to increase the training parameters, which can improve the model's performance. The Table \ref{tab:The number of learnable parameters across all tasks} below shows the proportion of parameters selected in our paper. The datasets include all tasks from FGVC, VTAB and GLUE. The Param represents the proportion of parameters updated using GRFT in a given task relative to the total model parameters. The Row/Column Number indicates the number of selected rows and columns. For most tasks in this paper, we only select no more than five rows/columns.

\end{document}